\documentclass{article}
\usepackage{ijcai17}

\usepackage{times}

\usepackage{helvet}
\usepackage{courier}
\usepackage{amsmath}
\usepackage{amssymb}
\usepackage{xcolor}
\usepackage{amsthm}
\usepackage{caption}
\usepackage{dsfont}
\usepackage{graphicx}
\usepackage{graphics}
\usepackage[boxed]{algorithm2e}
\usepackage{multirow}
\usepackage{xcolor,colortbl}
\usepackage{alltt}

\theoremstyle{definition}
\newtheorem{definition}{Definition}

\definecolor{Gray}{gray}{0.85}
\definecolor{LightCyan}{rgb}{0.88,1,1}

\newcommand{\crd}[0]{CUR$^2$LED }

\title{Clustering-Based Relational Unsupervised Representation Learning with an Explicit Distributed Representation}
\author{Sebastijan Duman\v{c}i\'{c} \and Hendrik Blockeel\\ 
Computer Science Department,
KU Leuven, Belgium  \\
\{sebastijan.dumancic,hendrik.blockeel\}@cs.kuleuven.be
}

\begin{document}

\maketitle

\begin{abstract}
The goal of unsupervised representation learning is to extract a new representation of data, such that solving many different tasks becomes easier.
Existing methods typically focus on vectorized data and offer little support for relational data, which additionally describe relationships among instances.
In this work we introduce an approach for \textit{relational} unsupervised representation learning.
Viewing a relational dataset as a hypergraph, new features are obtained by clustering vertices and hyperedges.
To find a representation suited for many relational learning tasks, a wide range of similarities between relational objects is considered, e.g. feature and structural similarities.
We experimentally evaluate the proposed approach and show that models learned on such latent representations perform better, have lower complexity, and outperform the existing approaches on classification tasks.
\end{abstract}

\section{Introduction}
\label{sec:Intro}
Every machine learning task inherently depends on the quality of provided features.
A good set of features is thus a crucial precondition for the good performance of any classifier. 
Yet, finding such a set in practice has proven to be a tedious and time-consuming task.
Furthermore, substantial domain knowledge and exploration are often required.
To address this issue,  \textit{representation learning} \cite{Bengio:2009} focuses on automatic learning of good multi-level data representations.

Representation learning methods include two categories.
Supervised representation learning learns a hierarchy of new features in a discriminative way. 
Thus, the representation is tailored specifically for a given task. 
An example of such an approach are convolutional neural networks.
In contrast, the unsupervised representation learning ($\mathcal{URL}$) \cite{Hinton504,Bengio07greedylayer-wise,RanzatoBL07} takes a \textit{generative stance}. 
Because it requires no supervision, such representation could be shared among multiple tasks. 
This is the direction we explore.
Intuitively, these methods find useful features by compressing the original data by means of a  multiple-clustering procedure, in which an instance can belong to more than one cluster.
The features obtained by $\mathcal{URL}$ typically indicate cluster assignments of each instance.
Consequently, a classifier learns from cluster memberships instead of the original data.

One major limitation of the existing methods is that they focus on vectorized data representations.
Hence, the ubiquitous and abundant structured and relational data are not well supported. 
In contrast to the vectorized representations, relational data describe instances together with their relationships. 
This is often viewed as a hypergraph\footnote{A hypergraph is a graph in which edges can connect more than two vertices.}, in which instances form vertices and their relationships form hyperedges.
Such data emerges in many real-life problems.
For instance, chemical and biological data describing molecules or protein interaction networks are naturally expressed in graph-structured formats.
Another example includes social networks, where many instances interact with each other.
A common language for expressing such data is \textit{predicate logic}, which further subsumes any data stored in  a relational database.

This work focuses on unsupervised representation learning with relational data.
To this end, we introduce \crd - a \textit{clustering-based unsupervised relational representation learning with explicit distributed representation}.
\crd is inspired by the work of \citeauthor{coates2011analysis}, \citeyear{coates2011analysis}, in which the authors introduce a general pipeline for learning a feature hierarchy by means of clustering.
Assuming a spatial order of features (i.e., pixels), the introduced framework (i) extracts image patches, i.e., subsets of pixel from the original images candidating as a potential high-level feature, (ii) pre-processes each patch (e.g. normalization and whitening), and (iii) learns a feature-mapping by clustering image patches.
The authors show that such general procedure with a simple k-means algorithm can perform as well as the specialized algorithms, such as auto-encoders and  Restricted Boltzmann machines.

\crd learns a new representation by clustering both instances and their relationships.
What is distinctive about \crd is that the relational structure is  preserved throughout the hierarchy, contrasted to the existing approaches that map relational structures onto a vectorized representation.
Another distinctive feature of \crd is the notion of \textit{similarity interpretation}.
When clustering relational data, a similarity of relational objects is an ambiguous concept.
Two relational objects might be similar according to their attributes, relationships, or a combination of both.
The notion of similarity interpretation precisely states the exact source of similarity used.
\crd exploits this ambiguity to its advantage by using the similarity interpretations to encode a \textit{distributed representation} of data.

Distributed representation is one of the pillars underlying the success of representation learning methods.
Intuitively, it refers to a concept of  reasonably-sized representation that captures a huge number of possible configurations \cite{Bengio2013RLR}.
In contrast to the one-hot representations which require $N$ parameters to represent $N$ regions,  distributed representations require $N$ parameters to represent up to $2^N$ regions.
The main difference is that a concept within a distributed representation is represented with several independently manipulated factors, instead of exactly one factor as with one-hot representations.
Thus, such representations are substantially more expressive. 
In that sense, the similarity interpretation defines the exact factors that can be manipulated individually to represent individual concepts.

The contributions of this paper include (i)  a general framework for learning relational feature hierarchies by means of clustering, (ii) a principled way of generating distributed relational representations based on different similarity interpretations, (iii) a general framework for hyperedge clustering and (iv) the experimental evaluation of the proposed framework.

In the following section, we briefly review related work.
Next, we outline our approach and discuss arising issues in Section \ref{sec:RL}.
We then briefly present the similarity measure used for clustering relational data, discuss its extension towards hyperedge clustering, and formally define the notion of similarity interpretation.
Experimental results are discussed in Section \ref{sec:Results}.

\section{Related work}
\label{sec:Related}

Clustering has been previously recognized as an effective way of enhancing relational learners.
\citeauthor{Popescul2004} (\citeyear{Popescul2004}) apply k-means clustering to the objects of each type in a domain, create predicates for new clusters and add them to the original data.
\textit{Multiple relational clustering (MRC)} \cite{Kok2007,Kok2008} is a relational probabilistic clustering framework based on Markov logic networks \cite{Richardson2006} clustering both vertices and relationships. 
Both approaches are instances of predicate invention tasks\cite{Kramer1995,Craven2001}, concerned with extending the vocabulary given to a learner by discovering novel concepts in data.
\crd differs in several ways.
Whereas \citeauthor{Popescul2004} develop a method specifically  for document classification, \crd is a general \textit{off-the-shelf} procedure that can be applied to any relational domain.
Moreover, CUR$^2$LED clusters both instances and relations, whereas \citeauthor{Popescul2004} cluster only instances.
In contrast to MRC which does not put any assumptions in the model, \crd is a more informed approach that explicitly defines different notions of relational similarity to be used for clustering.
Though MRC was used as a component in structure learning, it does not provide new language constructs, but simplifies the search over possible formulas.
\crd learns a model directly from the new features.

Much of the recent work focused on constructing the  \textit{embeddings} of relational constructs \cite{Niepert:NIPS2016,Bordes:2011:LSE,DBLP:journals/corr/YangYHGD14a,Bordes:2013:TEM,DBLP:journals/corr/HenaffBL15,DBLP:conf/icml/NiepertAK16}.
This work maps relational concepts to low dimensional vector spaces, and therefore loses the relationship information.
Moreover, it focuses on supervised learning.
Thus, \crd differs in that it learns a \textit{relational} feature hierarchy in an unsupervised manner.



\section{Representation learning via clustering}
\label{sec:RL}

Several issues arise from devising a general relational feature hierarchy pipeline, all of them due to the complexity of relational data.
Firstly, the issue is \textit{what should be clustered?}
In the i.i.d. case (drawn independently from the same population), the dataset contains only instances and their features, thus, one clusters the instances.
However, relational data additionally describes relationships among instances.
Furthermore, it can vary from a single large network of many interconnected entities (a \textit{mega-example}), to a set of many disconnected networks where each network is an example.
Ideally, one would address both cases.
\crd assumes that relational data is provided as a labelled hypergraph, where examples form vertices and relations between them form hyperedges, and does not make a distinction between the above-mentioned cases.
Formally said, the data structure is a typed, labelled hypergraph $H = (V,E, \tau, \lambda)$ with $V$ being a set of vertices, $E$ a set of hyperedges, $\tau$  a function  assigning a type to each vertex and hyperedge, and $\lambda$ a function assigning a vector of values to each vertex.

\begin{figure*}
    \centering
    \includegraphics[scale=0.232]{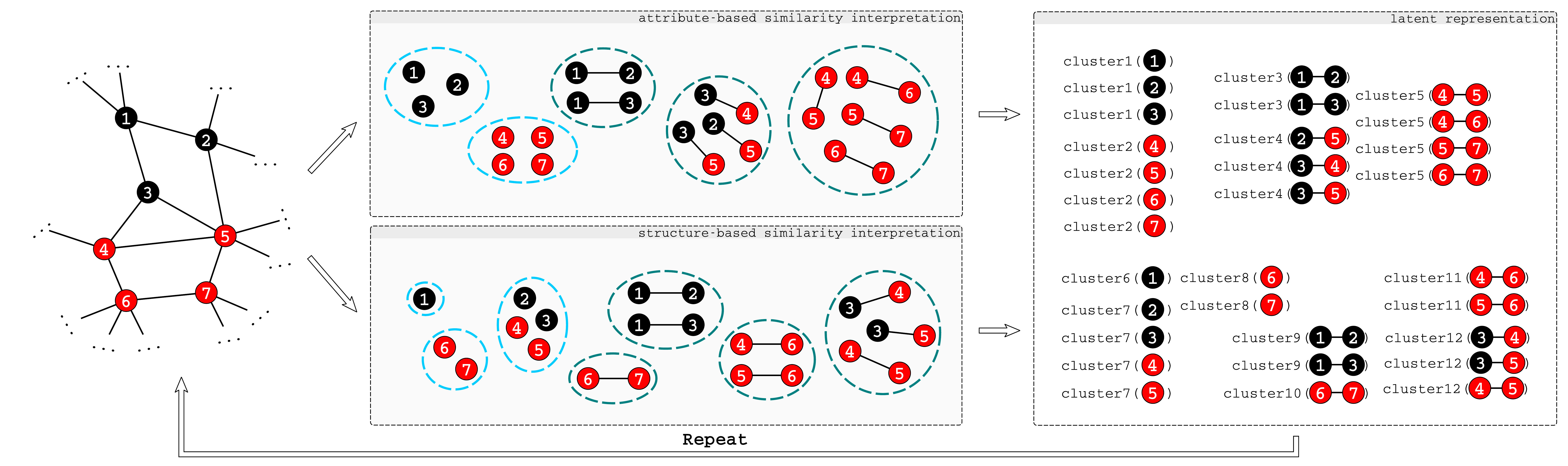}
    \caption{An illustration of \crd procedure. The left-most figure represents a given hypergraph, where colour of a vertex indicates its feature value. The graph (i.e., vertices and edges) is then clustered according to different similarity interpretations. The upper clustering is based on vertex attributes: the vertices are clustered into \textit{red} and \textit{black} ones, while the edges are clustered according to the colour of the vertices they connect. The bottom clustering is based on the structure of the neighbourhoods. The vertices are clustered into a group that have only \textit{black} neighbours (\{{\tt1}\}), only \textit{red} neighbours (\{{\tt 6,7}\}), and neighbours of both colours (\{{\tt 2,3,4,5}\}). The edges are clustered into a group of edges connecting \textit{black} vertices with only \textit{black} neighbours and \textit{black} vertices with \textit{red} neighbours (\{{\tt1-2,1-3}\}), a group of edges connecting \textit{red} vertices with only \textit{red} neighbours to \textit{red} vertices with neighbours of both colour (\{{\tt6-7}\}), and so on. The final step transforms the obtained clusterings into a relational representation. The procedure can further be repeated to create more layers of features. }
    \label{fig:curled}
\end{figure*}

\crd learns a new representation by clustering both \textit{vertices and hyperedges in a hypergraph}.
Considering that vertices have associated types, \crd does not allow mixing of types, i.e., a cluster can contain only vertices of the same type.
The same holds for hyperedges which connect vertices of different types.

The second issue emerges with \textit{the ambiguity of similarity in relational context.}
With the i.i.d. data, the features are the only source of similarity between examples.
In the relational context, a similarity can originate in the  features of relational objects, structures of their neighbourhoods (both features- and relationship-wise), interconnectivity or graph proximity, just to name a few.
Furthermore, which interpretation is needed for a particular task is not known in advance, making $\mathcal{URL}$ inherently more difficult.
Considering that \crd aims at finding a representation effective for many tasks, \crd addresses \textit{multiple interpretations of relational similarity simultaneously.}
How exactly that is achieved is discussed in the next section, together with a similarity measure used for this purpose.

The final issue concerns \textit{the structure of the feature hierarchy}.
Defining such hierarchy requires the specification of the number of layers, as well as the number of hidden features (i.e., clusters) within each layer.
How to automatically construct such hierarchies is currently under-explored.
Consequently, the performance of these methods is sensitive to the parameter setting, requiring substantial expertise in order to choose the optimal number of features.
This constitutes a major bottleneck for relational \textit{type-aware} feature hierarchies, as separate values should be chosen for each type in the data (and combination thereof for hyperedges).

To tackle this infeasibility, \crd builds upon a vast literature on \textit{the problem of clustering selection} \cite{Arbelaitz:2013}, which is concerned with the selection of optimal number of clusters from data.
Though the perfect method does not exist, automatic clustering selection mitigates the problem of the manual specification of feature hierarchies.
\crd leverages  two distinct approaches: \textit{(1) a difference-like criterion} \cite{Vendramin:2010}, and \textit{(2) a quality based criterion of Silhouette index} \cite{Rousseeuw:1987}.

Difference-like criteria assess relative improvements on some relevant characteristic of the data (e.g. within-cluster similarity) over a set of successive data partitions produced by gradually increasing the number of clusters ($N$).
It attempts to identify a prominent \textit{knee} - a point when the given quality measure \textit{saturates} and the further increase of $N$ can offer only marginal benefit.
Following the suggestion in \cite{Vendramin:2010}, we choose the number of clusters as the one that achieves the highest value of the following formula:

\begin{equation}
	D(k) = \left| \frac{C(k-1) - C(k)}{C(k) - C(k+1)} \right| - \alpha \cdot k
	\label{eq:Sat}
\end{equation}

where $C(k)$ is the intra-cluster similarity, $k$ is the number of clusters and $\alpha$ a user-specified penalty on the number of clusters.

The Silhouette index evaluates a \textit{cohesion}, i.e., how similar an instance is to its own cluster, and a \textit{separation}, i.e., how similar an instance is to the other clusters.
It is defined as:

\begin{equation}
    S(i) = \frac{a(i) - b(i)}{ \max \{ a(i), b(i)\}}
\end{equation}

where $i$ is an instance, $a(i)$ is an average dissimilarity of $i$ to the rest of the instances in the same cluster, and $b(i)$ the lowest dissimilarity of $i$ to any other cluster. 
Higher values indicate a better fit of the data.

\textbf{$\mathcal{C}$-representation.} 
Once the clusters are obtained, we will represent them in the following form.
For each cluster of vertices we create a unary predicate in the form of \texttt{cluster$ID$(vertex)} where {\tt vertex} is an identifier of a specific vertex.
Similarly, for each cluster of hyperedges we create a $n$-ary predicate in the form of \texttt{cluster$ID$(vertex$_1$,\ldots,vertex$_n$)}, which takes an ordered set of $n$ vertices as arguments.
We refer to the cluster-induced representation as a $\mathcal{C}$-representation.

The introduced pipeline is illustrated in Figure~\ref{fig:curled}.
\crd specified thus far describes a \textit{meta-procedure} how to use any clustering algorithm to obtain a latent representation.
In the experiments we use spectral and hierarchical clustering.

\section{Similarity of relational structures}
\label{sec:Clustering}

\crd relies on ReCeNT \cite{2016arXiv160408934D}, a relational clustering framework focused on clustering vertices in a hypergraph.
What makes ReCeNT an attractive relational clustering framework is the wide range of similarities it considers.
Furthermore, which similarity is used is easily adaptable with just a few parameters.
We provide a concise and intuitive description here, and refer the reader to the original paper for the details.

The core concept of ReCeNT is a \textit{neighbourhood tree} (NT).
The NT is a rooted directed graph describing a neighbourhood of a certain vertex in the hypergraph.
It provides a summary of all paths that can be taken, starting from that particular vertex.
The depth of a NT, i.e., how many vertices a path can contain excluding the root vertex, is pre-specified.
ReCeNT compares two vertices by comparing their NTs.

This comparison is achieved by first decomposing the NT into different multisets.
The multiset $V^l_{t}(g)$ contains all vertices of type $t$ at distance $l$ of a particular NT $g$.
$E^l(g)$ is the multiset of hyperedge labels between vertices of distances $l$ and $l+1$.
Finally, $B^l_{t,a}(g)$ is the multiset of values of attribute $a$ observed among the nodes of type $t$ at distance $l$.
Using only these three types of multisets, one can express a wide range of similarities.
What ReCeNT considers are:
\begin{enumerate}
    \item the similarity of the root vertices in terms of attribute values, by means of $B^0_{t,a}$ 
    \item the similarity of attribute values of the neighbouring vertices, by means of $B^{l>0}_{t,a}$ 
    \item the connectivity of the root vertices, by means of $V^l_t$ 
    \item the similarity of neighbourhoods in terms of the vertex identities, by means of $V^l_t$ 
    \item the similarity of hyperedge labels of two neighbourhoods, by means of $E^l$.
\end{enumerate}

Each of the components represents a distinct notion of similarity.
We will refer to them as \textit{core similarities}.
These core similarities can further be combined to represent more complex similarities.

\subsection{Hyperedge similarity}

In its original form, ReCeNT clusters vertices in a hypergraph.
However, \crd additionally clusters hyperedges.
Here we introduce a general framework for hyperedge clustering.
It views hyperedges as ordered sets of vertices, and thus ordered sets of NTs.

Let $\mathcal{N}$ be a set of NTs.
Let $\Theta$ denote summary operations on sets of values such as mean, minimum and maximum.
Let $\Lambda$ denote set operators such as union and intersection.
Let $f: \mathcal{N}^2 \rightarrow \mathbb{R}$ be a similarity between two NTs, e.g. the similarity measure introduced by ReCeNT.
The framework introduces two types of hyperedge similarity, namely \textit{combination} and \textit{merging}.

\begin{definition}
A \textit{combination similarity} is a function $c: \mathcal{N}^n \times \mathcal{N}^n \times \Theta \rightarrow \mathbb{R}$ which compares two hyperedges, $e_1 = (v_1^1,..., v_1^n)$ and $e_1 = (v_2^1,..., v_2^n)$, by comparing the individual NTs respecting the order, $s = \left(f(v_1^1,v_2^1),\ldots,f(v_1^i,v_2^i) \right)$, and summarizing respective similarities with $\theta \in \Theta$,  $\theta(s)$. 
\end{definition}

\begin{definition}
A \textit{merging similarity} is a function $m: 2^{\mathcal{N}} \times 2^{\mathcal{N}} \times \Lambda \rightarrow \mathbb{R}$ which compares two hyperedges, $e_1 = (v_1^1,..., v_1^n)$ and $e_1 = (v_2^1,..., v_2^n)$, by first merging the NTs within a hyperedge with merging operator $\lambda \in \Lambda$, $s_1 = \lambda (v_1^1,\ldots,v_1^i)$ and $s_2 = \lambda (v_2^1,\ldots,v_2^i)$ and comparing the resulting NTs, $f(s_1,s_2)$.  
\end{definition}

Merging two NTs involves merging their respecting multisets with a merging operator $\lambda$, respecting the level.
For instance, consider \textit{set union} as $\lambda$, and $g'$ and $g''$ as the NTs to be merged.
Then, merging the multisets $V^l_{t}(g')$ and $V^l_{t}(g'')$ results in a multiset $V^l_{t}\left(\lambda(g',g'')\right) = V^l_{t}(g') \cup V^l_{t}(g'')$.

Both formulations reduce the problem to the comparison of NTs, but offer alternative views.
While \textit{merging} ignores the order of vertices in a hyperedge, \textit{combination} respects it.
Accordingly, \textit{merging} describes the neighbourhood of a hyperedge, while \textit{combination} examines the similarity of vertices participating in a hyperedge.
In this work we use \textit{union} as the merging operator, and \textit{mean} as the combination operator.

\subsection{Similarity interpretation}

Finally, we formally introduce the notion of similarity interpretation.

\begin{definition}
Let $(w_1,w_2,w_3,w_4,w_5)$ be the weights associated with the \textit{core similarities}.
A \textit{similarity interpretation} is the value assignments to the weights $(w_1,w_2,w_3,w_4,w_5)$. 
\end{definition}

Thus, it allows us to precisely control aspects of similarity considered for representation learning.
For example, setting $w_1 = 1, w_{2,3,4,5} = 0$ uses only the attributes of vertices for comparison.
Setting $w_3 = 1, w_{1,2,4,5} = 0$ on the other hand would identify clusters as a densely connected components. 
As the similarity interpretation is provided by the user, we say it \textit{explicitly} defines the distributed representation.

\section{Experiments and results}
\label{sec:Results}

\begin{table*}
\captionsetup{justification=centerlast}
    \begin{center}
        \footnotesize
        \caption{Performance comparison for TILDE models learned on the original and $\mathcal{C}$-representations. The first column specifies the parameters used for $\mathcal{C}$-representation, i.e., clustering algorithm (S-spectral, H-hierarchical), selection criterion and its parameter values. Both accuracies on a test set (Acc) and complexities (Cplx) are reported.}
        \resizebox{0.9\textwidth}{!}{%
        \begin{tabular}[th]{c|c|c|c|c|c||c|c|c|c|c|c|c|c|}
            \cline{2-14}
                \multirow{2}{*}{}& \multirow{2}{*}{\textbf{Setup}} & \multicolumn{2}{|c|}{\textbf{IMDB}} & \multicolumn{2}{|c||}{\textbf{UWCSE}} & \multicolumn{2}{|c|}{\textbf{Mutagenesis}} & \multicolumn{2}{|c|}{\textbf{Terrorists}} & \multicolumn{2}{|c|}{\textbf{Hepatitis}} & \multicolumn{2}{|c|}{\textbf{WebKB}} \\
                                 &                                 &  Acc & Cplx &  Acc & Cplx & Acc & Cplx & Acc & Cplx & Acc & Cplx & Acc & Cplx \\
                \cline{2-14}
                \cline{2-14}
                                 & \textbf{Original}                                                    & 1.0  & 2.0        & 0.99 & 3.0                & 0.76 & 27.2                   & \textbf{0.72} & 86.4       & 0.81 & 22.4       & 0.81 & 18.2 \\
                \hline
                \hline
       \parbox[t]{2mm}{\multirow{8}{*}{\rotatebox[origin=c]{90}{\textbf{merging}}}} & S,$\alpha=0.01$   & 1.0  & 1.0        & 0.99 & 1.2                 & 0.79 & 6.6                   & \textbf{0.71} & 34.4       & 0.86 & 19.66             & \textbf{0.89 }& 13.6  \\
                \cline{2-14}
                                                                                    & S,$\alpha=0.05$   & 1.0  & 1.0        & 0.99 & \textbf{1.0}        & 0.78 & 2.4                   & 0.65 & 21.6                & 0.90 & 7.6                & 0.85 & 15.6 \\
                \cline{2-14}
                                                                                    & S,$\alpha=0.1$    & 1.0  & 1.0        & 0.99 & 1.2                 & 0.78 & \textbf{1.8}          & 0.66 & 32.4                & 0.90 & 6.5               & \textbf{0.87} & 17.8  \\
                \cline{2-14}
                                                                                    & S,silhouette      & 1.0  & 1.0        & 0.99 & \textbf{1.0}        & 0.78 & \textbf{2.0}          & 0.6 & 23.6                 & \textbf{0.93} & \textbf{5.33}       & \textbf{0.87} & 14.8 \\
                \cline{2-14}                                                       
                                                                                    & H,$\alpha=0.01$   & 1.0  & 1.0        & 0.98 & 4.4                 & \textbf{0.83} & \textbf{2.0} & 0.48 & \textbf{9.4}        & 0.86 & 12.0       & 0.83 & 12.6  \\
                \cline{2-14}
                                                                                    & H,$\alpha=0.05$   & 1.0  & 1.0        & 0.99 & 4.2                 & \textbf{0.83} & \textbf{2.0} & 0.48 & 11.6                & 0.82 & 16.0       & 0.69 & 27.2 \\
                \cline{2-14}
                                                                                    & H,$\alpha=0.1$    & 1.0  & 1.0        & 0.99 & 4.0                 & 0.79 & 5.2                   & 0.47 & \textbf{8.8}        & 0.82 & 13.4       & 0.61 & 32.2 \\
                \cline{2-14}
                                                                                    & H,silhouette      & 1.0  & 1.0        & 0.98 & \textbf{1.0}        & 0.80 & 3.4                   & 0.47 & 13.0                & \textbf{0.93} & 8.66       & 0.68 & 18.0 \\
                \hline
                \hline
   \parbox[t]{2mm}{\multirow{8}{*}{\rotatebox[origin=c]{90}{\textbf{combination}}}} & S,$\alpha=0.01$   & 1.0  & 1.0        & 0.99 & 1.2                 & 0.79 & \textbf{2.0}          & \textbf{0.72} & 24.0       & 0.90 & 7.6        & \textbf{0.91} & 11.8 \\
                \cline{2-14}
                                                                                    & S,$\alpha=0.05$   & 1.0  & 1.0        & 0.99 & \textbf{1.0}        & 0.79 & \textbf{2.0}          & 0.69 & 22.8                & 0.88 & 12.2       & 0.86 & \textbf{10.0} \\
                \cline{2-14}
                                                                                    & S,$\alpha=0.1$    & 1.0  & 1.0        & 1.0 & \textbf{1.0}         & 0.76 & \textbf{2.0}          & 0.66 & 16.8                & 0.90 & 12.6       & 0.87 & 17.0 \\
                \cline{2-14}
                                                                                    & S,silhouette      & 1.0  & 1.0        & 0.99 & \textbf{1.0}        & 0.77 & \textbf{2.0}          & 0.6  & 24.2                & \textbf{0.93} & 16.4       & \textbf{0.88} & 13.8\\
                \cline{2-14}
                                                                                    & H,$\alpha=0.01$   & 1.0  & 1.0        & 0.99 & 2.8                 & 0.79 & 4.0                   & 0.51 & 30.6                & 0.80 & 29.33      & 0.83 & 12.6 \\
                \cline{2-14}
                                                                                    & H,$\alpha=0.05$   & 1.0  & 1.0        & 0.99 & 2.8                 & 0.78 & 2.8                   & 0.51 & 30.6                & 0.82 & 16.33      & 0.69 & 27.2 \\
                \cline{2-14}
                                                                                    & H,$\alpha=0.1$    & 1.0  & 1.0        & 0.99 & 2.8                 & 0.78 & 11.0                  & 0.50 & 27.3                & 0.78 & 14.0       & 0.61 & 32.2   \\
                \cline{2-14}
                                                                                    & H,silhouette      & 1.0  & 1.0        & 0.99 & 2.0                 & 0.80 & 4.0                   & 0.50 & 30.0                & 0.83 & 11.6       & 0.68 & 18.0  \\
                \hline
                \hline
        \parbox[t]{2mm}{\multirow{3}{*}{\rotatebox[origin=c]{90}{\textbf{MRC}}}}    & $\lambda=-1$      & 1.0  & 1.0       & 0.93 & 21.0        & 0.6  & 0          & 0.64 & 138.7      &  0.61 & 99.4        &0.64 & 44.4 \\
                \cline{2-14}
                                                                                    & $\lambda=-5$      & 1.0  & 1.0       & 0.95 & 25.9        & 0.63  & 23.5      & 0.50 & 126.5        & 0.84 & 64.8       &0.68 & 40.0 \\
                \cline{2-14}
                                                                                    & $\lambda=-10$     & 1.0  & 1.0       & 0.96 & 13.7        & 0.72  & 35.0      & 0.51 & 102.1        & 0.57 & 5.7         &0.66 & 40.8 \\
                \cline{1-14}

        \end{tabular}
        }
        \label{tab:Results}
    \end{center}
\end{table*}

\textbf{Datasets.}
We have used the following 6 datasets to evaluate the potential of this approach.
The IMDB dataset describes a set of movies with people acting in or directing them.
The UW-CSE dataset describes the interactions of employees at the University of Washington and their roles, publications and the courses they teach.
The Mutagenesis dataset describes chemical compounds and atoms they consist of. 
The WebKB dataset consists of pages and links collected from the Cornell University's web page.
The Terrorists dataset describes terrorist attacks each assigned one of 6 labels indicating the type of the attack.
The Hepatitis dataset describes a set of patients with hepatitis types B and C.
\vspace{1pt}

\textbf{Evaluation procedure.}
In principle, a latent representation should make learning easier by capturing complex dependencies in data more explicitly.
Though that is difficult to formalize, a consequence should be that a model learned on the latent representation is (i) \textit{less complex}, and (ii) possibly \textit{performs better}.
To verify whether that is the case with the representation created by \crd, we answer the following questions:
\begin{itemize}
	\setlength\itemsep{0.06em}
    \item[\textbf{(Q1)}] \textit{do representations learned by \crd induce models of lower complexity compared to the ones induced on the original representation?}
    \item[\textbf{(Q2)}] \textit{if the original data representation is sufficient to solve a task efficiently, does $\mathcal{C}$-representation preserves the relevant information?}
    \item[\textbf{(Q3)}] \textit{if the original data representation is not sufficient to solve the task, does a $\mathcal{C}$-representation improve the performance of a relational classifier?}
    \item[\textbf{(Q4)}] \textit{can the appropriate parameters for a specific dataset be found by the model selection?}
    \item[\textbf{(Q5)}] \textit{how does \crd compare to MRC, which is the closest related work?}
\end{itemize}

In order to do so, we use TILDE \cite{Blockeel1998285}, a relational decision tree learner, and perform 5-fold cross validation. 
$\mathcal{C}$-representations and TILDE were learned on training folds, and the objects from the test fold were mapped to the $\mathcal{C}$-representation and used to test TILDE.
The following similarity interpretation were used for each dataset: (0.5,0.5,0.0,0.0,0.0), (0.0,0.0,0.33,0.33,0.34), (0.2,0.2,0.2,0.2,0.2).
The first set of weights uses only the attribute information, the second one only the link information, while the last one combines every component.

As a complexity measure of a model we use the number of nodes a trained TILDE model has.
We use the following values for the $\alpha$ parameter in Equation~\ref{eq:Sat}: $\{ 0.1,0.05, 0.01 \}$.
In the case of MRC, we used the following values for the $\lambda$ parameter: $\{-1,-5,-10\}$.
The $\lambda$ parameter has the same role as $\alpha$ in the proposed approach, affecting the number of clusters chosen for each type\footnote{Note that it is difficult to exactly match the values of $\alpha$ and $\lambda$ as both methods operate on different scales, and the authors do not provide a way how to choose an appropriate value}.

\subsection*{Results}

To answer the above mentioned questions, we perform two types of experiments.
Table~\ref{tab:Results} summarizes the results of cross validation.
The accuracies on test set and the complexities of TILDE models are stated for both original and $\mathcal{C}$-representations.
Table~\ref{tab:MS} summarizes the results of the model selection where we dedicate one fold as a \textit{validation set}, and perform the cross validation on the remaining folds to identify the best parameter values (i.e., the choice of a clustering algorithm, a clustering selection procedure and the appropriate hyperedge similarity) for each dataset.

\textbf{Q1.}
Table~\ref{tab:Results} shows that the models learned on $\mathfrak{C}$-representation consistently have lower complexity than the ones learned on the original data.
That is especially the case when $\mathcal{C}$-representation is obtained by spectral clustering, which consistently results in a model of a lower complexity.
The reduction of complexity can even be surprisingly substantial, for instance on the Mutagenesis and Hepatitis datasets where the model complexities are reduced by factors of 10 and 4, respectively.
When the $\mathcal{C}$-representation is obtained with hierarchical clustering, models of lower complexity are obtained on all datasets except the WebKB and UWCSE datasets.
These results suggest that the $\mathcal{C}$-representation in general makes complex dependencies easier to detect and express.

\textbf{Q2.}
The IMDB and UWCSE datasets are considered as \textit{easy} relational datasets, where the classes are separable by a single attribute or a relationship.
Thus, TILDE is able to achieve almost perfect performance  with the original data.
The original representation is therefore sufficient to solve the task, and we are interested whether  the relevant information will be preserved within the $\mathcal{C}$-representation.
The results in Table~\ref{tab:Results} do suggest so, as TILDE achieves identical performance regardless of the representation.

\textbf{Q3.}
The remaining datasets are more difficult than the previously discussed ones.
On the Mutagenesis, Hepatitis and WebKB datasets, $\mathcal{C}$-representation improves the performance.
On the Terrorists dataset, however, no improvement in performance is observed.
What distinguishes this dataset from the others is that it contains only two edge types (indicating co-located attack, or ones organized by the same organization), an abundant number of features, while other datasets are substantially more interconnected.
Thus, focusing on the relational information is not as beneficial as the features themselves. 

These results suggest that $\mathcal{C}$-representations indeed improve performance of the classifier, compared to the one learned on the original data representation.
First, the $\mathcal{C}$-representation created with spectral clustering consistently performs better on all datasets, except the Terrorists one.
Second, if the learning  task does not have a strong relational component, then $\mathcal{C}$-representations are not beneficial and can even hurt the performance.
Third, the choice of a clustering algorithm matters, and spectral clustering does a better job in our experiments - it always results in improved or at least equally good performance.
Fourth, the choice of treating hyperedges as ordered (by \textit{combination}) or unordered (\textit{merging}) sets is data-dependent, and the difference in performance is observed.

Combining the results from \textbf{Q1}, \textbf{Q2} and \textbf{Q3} shows that \textit{the main benefit} of \crd is \textit{the transformation of data such that it becomes easier to express complex dependencies}.
Consequently, the obtained models have lower complexities and  their performance often improves.

\textbf{Q4.} To ensure that the previously discussed results do not over-fit the data, we additionally perform model selection.
We dedicate one fold as the \textit{validation set}, and use the remaining folds to find the best parameter values of both \crd and TILDE.
Table \ref{tab:MS} summarizes the results and reports the selected choice of parameter, together with the performance on the validation set.
These results are consistent with the ones in Table~\ref{tab:Results}: $\mathcal{C}$-representation improves the performance in the majority of cases, and the selected parameters correspond to the best performing ones in Table~\ref{tab:Results}.

\begin{table}
\captionsetup{justification=centerlast}
\begin{center}
	\caption{Model selection results. For each dataset, a selected parameters are reported together with the accuracies on the training and test sets. The first element indicates the selected clustering algorithm (S-spectral, H-hierarchical), the second one the clustering selection criteria, while the last one indicates the hyperedge similarity (C-combination, M-merging). The last column indicates the performance on the original data representation. }
    \resizebox{0.98\linewidth}{!}{%
    	\begin{tabular}[t]{|c|c|c|c|c|}
        \hline
        \textbf{Dataset} & \textbf{Parameters} & \textbf{Training} & \textbf{Validation} & \textbf{Original}\\
        \hline
        \hline
        \textbf{IMDB} & all & 1.0 & \textbf{1.0} & \textbf{1.0}\\
        \hline
        \textbf{UWCSE} & S, silhouette, C & 0.99 & \textbf{1.0} & 0.99 \\
        \hline
        \textbf{Mutagenesis} & H, $\alpha$=0.01, M & 0.86 & \textbf{0.84} & 0.79 \\
        \hline
        \textbf{Hepatitis} & S, silhouette, M & 0.92 & \textbf{0.89} & 0.8  \\
        \hline
        \textbf{WebKB} & S, $\alpha$=0.01, C & 0.88 & \textbf{0.88} & 0.79 \\
        \hline
        \textbf{Terrorists} & S,$\alpha$=0.01, C & 0.70 & 0.69 & \textbf{0.71} \\
        \hline
        \end{tabular}
    
    }
    \label{tab:MS}

\end{center}

\end{table}

Considering the computational cost, \crd is an expensive step which depends on the size of a dataset like all representation learning approaches.
Performing a 5-fold cross validation on a single CPU took approximately 5 minutes for the IMDB and UWCSE datasets, 24 hours for the Terrorists dataset and approximately a week for the remaining datasets.
Though expensive, latent representation has to be created only once and can be reused for many tasks with the same dataset.
Moreover, \crd is easily parallelizable which can substantially improve its efficiency.

\textbf{Q5.}
Table~\ref{tab:Results} shows that \crd substantially outperforms MRC on all datasets, achieving better performance on all datasets except the IMDB.
Moreover, MRC rarely shows benefit over the original data representation, with an exception on the Hepatitis dataset.
Considering the model complexity, the models learned on MRC-induced representation are substantially more complex than the ones learned on $\mathcal{C}$-representations.
Table~\ref{tab:Size} summarize the number of clusters created by \crd and MRC.
One can see that MRC creates substantially more clusters than \crd. 
Because of this, the found clusters contain only a few objects which makes it difficult to generalize well, and increases the model complexity.
The number of clusters found by \crd is relatively high, because it finds a representation of data suitable for many classification tasks over the same datasets.
Thus, most of the features are redundant for one specific task, but clearly contain better information as the models learned on them perform better and have lower complexity.

\begin{table}[t]
\captionsetup{justification=centerlast}
\begin{center}
\footnotesize
\caption{Vocabulary sizes. M indicates MRC, while S and H indicate \crd representations with spectral and hierarchical clustering, respectively. Vocabulary sizes obtained with \textit{merging} and \textit{combination} similarities were similar, so only the one for merging is reported. }
\resizebox{0.98\linewidth}{!}{%
\begin{tabular}[t]{|c|c|c|c|c|c|c|}
	\hline
	\textbf{Setup} & \textbf{UW} & \textbf{Muta} &\textbf{WebKB} & \textbf{Terror}  & \textbf{IMDB} & \textbf{Hepa} \\
	\hline
	\textbf{Original} 				& 10         & 12  	     & 775     	& 107  		& 5  & 22 \\
	\hline
	\hline
	\textbf{S}, $\alpha=0.01$ 		& 109         & 53   	 & 65 		& 30	 	& 75  & 85 \\
	\hline
	\textbf{S}, $\alpha=0.05$ 		& 87      	  & 37       & 63 		& 26 		& 69  & 66 \\
	\hline
	\textbf{S}, $\alpha=0.1$ 		& 72          & 31  	 & 57  		& 24 		& 59  & 28  \\
    \hline
    \textbf{S}, silhouette 			& 93          & 17  	 & 59  		& 37 		& 74  & 79  \\
	\hline
	\hline
	\textbf{H}, $\alpha$=$0.01$ 	& 93          & 38   	 & 64   	& 25  		& 69  & 62 \\
	\hline
	\textbf{H}, $\alpha$=$0.05$ 	& 85           & 34   	 & 64  		& 20  		& 65  & 50 \\
	\hline
	\textbf{H}, $\alpha=0.1$ 		& 68           & 22   	 & 58  		& 18   		& 55  &  46 \\
    \hline
    \textbf{H}, silhouette  		& 85           & 20  	 & 55  		& 43   		& 64  & 61 \\
	\hline
	\hline
	\textbf{M}, $\lambda$=$-1$ 		& 183          & 535     & 817		& 318  		& 49   &  655	   \\
	\hline
	\textbf{M}, $\lambda$=$-5$ 		& 140          & 346   	 & 331		& 116    	& 38   & 297	   \\
	\hline
	\textbf{M}, $\lambda$=$-10$		& 49          & 224 	 & 219		& 91  		& 18   & 120	   \\
	\hline
	
\end{tabular} 
}

\label{tab:Size}

\end{center}
\end{table}

\section{Conclusion}
\label{sec:Conc}

This work introduces \crd - a clustering-based framework for unsupervised representation learning with relational data, which describes both instances and relationships between them.
Viewing relational data as hypergraph, \crd learns new features by clustering both instances and their relationships. i.e., vertices and hyperedges in the corresponding hypergraph.
To support such procedure, we introduce a general hyperedges clustering framework based on similarity of vertices participating in the hyperedge.
A distinct feature of \crd is the way it uses the ambiguity of similarity within relational data, i.e., whether two relational objects are similar due to their features of relationships, to generate distributed representation of data.
We design several experiments to verify the usefulness of latent representation generated by \crd.
The results show that the latent representations created by \crd provide a better representation of data that results in models of lower complexity and better performance. 
In future work, we will extend \crd towards semi-supervised settings, and investigate alternative ways for learning a distributed representations directly from data.

\bibliographystyle{named}
\bibliography{ijcai17}

\end{document}